\documentclass[]{interact}
\usepackage{multirow}
\usepackage{makecell}
\usepackage{xcolor}
\usepackage{algorithm}
\usepackage{epstopdf}% To incorporate .eps illustrations using PDFLaTeX, etc.
\usepackage[caption=false]{subfig}% Support for small, `sub' figures and tables
\usepackage[natbibapa,nodoi]{apacite}% Citation support using apacite.sty. Commands using natbib.sty MUST be deactivated first!

\theoremstyle{plain}% Theorem-like structures provided by amsthm.sty

\theoremstyle{definition}

\theoremstyle{remark}

\begin{document}

%\articletype{ARTICLE TEMPLATE}% Specify the article type or omit as appropriate

\title{Overlapping Word Removal is All You Need: Revisiting Data Imbalance in Hope Speech Detection}

%\author{\textbf{Anonymous Submission}}
\author{
\name{Hariharan RamakrishnaIyer LekshmiAmmal \textsuperscript{a}\thanks{CONTACT Hariharan R L. Email: hariharanrl.197it003@nitk.edu.in},Manikandan Ravikiran \textsuperscript{b},Gayathri Nisha \textsuperscript{a},Navyasree Balamuralidhar \textsuperscript{a},Adithya Madhusoodanan\textsuperscript{a},Anand Kumar Madasamy\textsuperscript{a},Bharathi Raja Chakravarthi\textsuperscript{c}}
\affil{\textsuperscript{a} Department of Information Technology, National Institute of Technology, Karnataka Surathkal, India}
\affil{\textsuperscript{b}Georgia Institute of technology, Atlanta, Georgia}
\affil{\textsuperscript{c}Insight SFI Research Centre for Data Analytics, Data Science Institute, National University of Ireland Galway, Galway, Ireland}
}

\maketitle

\begin{abstract}
Hope Speech Detection, a task of recognizing positive expressions, has made significant strides recently. However, much of the current works focus on model development without considering the issue of inherent imbalance in the data. Our work revisits this issue in hope-speech detection by introducing focal loss, data augmentation, and pre-processing strategies.  Accordingly, we find that introducing focal loss as part of Multilingual-BERT's (M-BERT) training process mitigates the effect of class imbalance and improves overall F1-Macro by 0.11. At the same time, contextual and back-translation-based word augmentation with M-BERT improves results by 0.10 over baseline despite imbalance. Finally, we show that overlapping word removal based on pre-processing, though simple, improves F1-Macro by 0.28. In due process, we present detailed studies depicting various behaviors of each of these strategies and summarize key findings from our empirical results for those interested in getting the most out of M-BERT for hope speech detection under real-world conditions of data imbalance. Our codes and models are made available on GitHub \footnote[2]{\url{https://github.com/hariharanrl/lre\_hope\_2021}}.
\end{abstract}


\section{Introduction}\label{intro}

Hope Speech Detection is the task of identifying expressions that is positive, encouraging, supportive, and inspiring promise of the future \citep{chakravarthi-2020-hopeedi}. One such example of Hope and Non-Hope speech is shown below.
\\
\\
\fbox{%
\centering
    \parbox{4.5in}{%
       \textbf{Hope Speech:} I'm so proud for her.\\
       \textbf{Non-Hope Speech:} Yup it’s definitely in your blood...your evil demonic blood...shes pure satanist wake up ppl.
    }}
\\

With growing social media applications identifying hope speech is more vital, especially in encouraging marginalized communities such as people from Lesbian, Gay, Bisexual, Transgender, and Queer (LGBTQ) communities, racial minorities, women in the fields of Science, Technology, Engineering, and Management (STEM) and people with disabilities \citep{DBLP:conf/emnlp/WangJ18}. Moreover, with  84\% of social media traffic generated by teenagers \citep{auxier_anderson_2021}, enforcing positivity is indispensable. Besides, due to the natural state of social media with the presence of online harassment, violent content, and abuse, inspiring people with positive comments have become an imperative need across social media platforms. Further, showing positive content tends to reflect more positive behavior by the people in social media \citep{Kramer8788}. Besides, with the recent pandemic resulting in a lack of social interaction and depression, there is a need to reinforce positivity and hope. Considering these needs, coupled with the availability of sizeable pretrained language models and release of open datasets, research on hope speech detection has seen meaningful traction where many of the works exhibit remarkable accuracy gains \citep{chakravarthi-2020-hopeedi}. However, despite such recent success, much of the study on hope speech detection is mainly directed towards fine-tuning large language models without examining the following inherent issues.

\begin{itemize}
    \item \textbf{Data Imbalance}: Firstly, due to the natural process of data collection and the current natural skew of social media users towards writing Non-Hope Speech, the dataset is highly imbalanced (See Table \ref{tab:datadistrib}). Meanwhile, Deep Neural Networks (DNN's), such as BERT and its alternatives which are used currently, often encounter generalization issues under conditions of data imbalance \citep{DBLP:journals/pami/DongGZ19}. Unfortunately, such imbalance problems are ignored by current research works on hope speech detection. More specifically, a survey of existing literature exhibits that out of 22 papers, 16 of them use a variant of BERT \citep{DBLP:conf/naacl/DevlinCLT19}, 3 of them employ simpler DNN architectures with word embeddings, with all of them not accounting for the data imbalance issue. 
    
\begin{table}[!htb]
\tbl{Training Data Distribution (\%) in Hope Speech Detection Dataset.}
{ \begin{tabular}{cccc}
        \toprule
        Language & Hope & Non-Hope & Other \\ \midrule
        English & 8.61 & 91.28 & 0.11 \\ \midrule
        \end{tabular}}
\label{tab:datadistrib}
\end{table}

    \item \textbf{Word Overlap Issue}: Secondly, to the best of our knowledge, none of the existing works consider the issue of word overlap between the Hope and Non-Hope speech comments. Our analysis reveals that out of 33458 words in the training vocabulary of the hope speech detection dataset, 5084 words are present in both  Hope and Non-Hope classes. Such overlap may lead to over-estimating models' performance, effectively rendering much of the current benchmarks useless. Moreover, such an issue often leads to bias in model and parameter selection towards overfitting, impacting model generalization. Furthermore, the issue is more prevalent when combined with the problem of data imbalance. 
    \item \textbf{Overlooked Preprocessing Stages}: Also, much of the research on hope speech detection overlooks pre-processing stages, with none of them explicitly mapping the relationship between data pre-processing techniques and their effect on imbalanced data distributions like the one seen in the hope speech detection problem. Specifically, our survey of existing works reveals that only two of the studies focus on some form of pre-processing..
    \item \textbf{Inaccurate Accuracy Measure:}  To date, all of the works on hope speech detection employ a weighted average F1 score. While this measure is reasonable, it is unsuitable for practical applications where the less expressed Hope Speech class does not have equal importance like Non-Hope class, making it more unreliable when evaluating the model performance, considering the aim is to develop a model that performs well on all classes, including minority. %While this measure is reasonable, it is unsuitable for practical applications where the underrepresented Hope Speech class does not have equal importance like Non-Hope class and is more unreliable when it comes to evaluating model performance, considering the aim is to develop a model that performs well on all the classes, including the minority classes.

\end{itemize}

Accordingly, we revisit the problem of hope speech detection by making the following contributions.

\begin{itemize}
    \item Firstly, we introduce a baseline benchmark for M-BERT on the English hope speech detection dataset. To this end, we report macro averaged results suitable for practical applications and empirically verify previously mentioned issues. 
    \item Secondly, to account for data imbalance in the context of hope speech detection, we study (a) Focal loss \citep{DBLP:journals/pami/LinGGHD20} instead of cross-entropy loss to train M-BERT,  (b) Contextual \citep{DBLP:conf/naacl/Kobayashi18} and Back-Translation \citep{DBLP:conf/acl/SennrichHB16} data augmentation to mitigate data imbalance. While the former reshapes the standard cross-entropy loss by penalizing the loss assigned to well-classified instances, accounting for imbalance, the latter generates new samples using existing samples from the Hope class. To this end, we find (a) focal loss when coupled with M-BERT improves F1-Macro by 0.1126 and 0.1173\% and (b) data augmentation improves F1-Macro by a maximum of 0.1131 and 0.1043  on validation and test splits of hope speech detection dataset.
    \item Finally, to account for word overlap issues, we propose a simplistic word removal algorithm as a preprocessing step, which focuses on removing conflicting context, improving F1-Macro  by 0.2786 and 0.2828, respectively, on validation and test sets.
    
\end{itemize}

The rest of the paper is organized as follows. First, in section \ref{background_prelim}, we briefly review the literature on hope speech detection, imbalanced text classification,  preprocessing, and data augmentation strategies. Followed by dataset and experimental setup are explained in section \ref{experimental_setup}. Then, in section \ref{tricks}, we present focal loss, data augmentation strategy, and word removal preprocessing algorithm post which the baseline results and other experiments are presented in section \ref{results}.  Finally, we conclude with hints on possible future works in section \ref{conclusion}.

\section{Related Work}\label{background_prelim}

In this section, we present literature on approaches for hope speech detection (section \ref{s1}), handling data imbalance (section \ref{s2}) and preprocessing approaches to improve results (section \ref{s3}).

\subsection{Approaches for Hope Speech Detection}\label{s1}

The currently published approaches for hope speech detection could be broadly divided into three categories as explained.
\begin{itemize}

    \item \textbf{Classical Approaches:} Classical approaches have seen very few works in hope speech detection, with notable of them by \citep{dave-etal-2021-irnlp-daiict} using TF-IDF character n-grams coupled with pre-trained MuRIL embeddings \citep{DBLP:journals/corr/abs-2103-10730}  for text representation and Logistic Regression and Linear SVM for classification.  
    \item \textbf{Hybrid Approaches:}  Hybrid approaches typically employ some form of deep learning architecture involving Convolutional Neural Network (CNN) or Recurrent Neural Network (RNN). These are coupled with one or more variants of embedding representations. In this line, there are works by \citep{m-k-a-p-2021-ku} using RNN with context-aware string embeddings and pooled document embeddings for word representations. Following this, there is work by \citep{balouchzahi-etal-2021-mucs-lt} which revealed three models, namely, CoHope-ML, CoHope-NN, and CoHope-TL based on an ensemble of classifiers, neural network (NN), and Bi-directional Long Short Term Memory (Bi-LSTM) with 1-D CNN model. Moreover, there are more simplistic variants of CNN-LSTM by \citep{saumya-mishra-2021-iiit} which uses GloVe \citep{DBLP:conf/emnlp/PenningtonSM14} and Word2Vec \citep{DBLP:journals/corr/abs-1301-3781} embeddings.
    \item \textbf{Language Modelling}: Among the methods in the literature, language models seem to enjoy a special place due to their wide usage and ease of training due to open source tools. Almost all the works use language models being used for hope speech detection, beginning with an exhaustive benchmark by \citep{puranik-etal-2021-iiitt} using variants of BERT. There are also individual works on using ALBERT \citep{DBLP:conf/slt/ChiCWHC0L21}, DistilBERT \citep{DBLP:journals/corr/abs-1910-01108}, XLM-RoBERTa \citep{DBLP:conf/acl/ConneauKGCWGGOZ20}, MuRIL \citep{DBLP:journals/corr/abs-2103-10730} to classify dataset for English, Malayalam and Tamil languages. Alternatively, there are a subset of works that employ M-BERT directly \citep{puranik-etal-2021-iiitt,s-etal-2021-ssn} and then few which use it for representation learning \citep{ziehe-etal-2021-gcdh,mahajan-etal-2021-teamuncc,que-2021-simon-lt}. Moreover, some works coupled Language models with classification approaches like \citep{huang-bai-2021-team} which consolidated the XLM-RoBERTa language model and the TF-IDF. Finally, there are few works \citep{awatramani-2021-hopeful,sharma-arora-2021-spartans} that employ some form of preprocessing and augmentation using back-transliteration of the comments. However, they do not apply to the English Language.
\end{itemize}

Overall much of the current works focus on using language models either directly or in the form of feature extractors without considering the issue of data imbalance. In our work, we study M-BERT by explicitly focusing on the impact of data imbalance, and in the due process to accommodate data imbalance, we propose training M-BERT with Focal Loss.

\subsection{Approaches for Handling Data Imbalance}\label{s2}

Handling data-imbalance is done either through data-level methods and algorithm-level methods with former create more data to balance the minority classes and later proposing algorithmic modifications. Fundamental of data level technique is Synthetic Minority Oversampling Technique (SMOTE) \citep{DBLP:journals/jair/ChawlaBHK02} and its variants  SMOTE-SVM \citep{DBLP:journals/ijkesdp/NguyenCK11}. Many more methods exists, including usage of synonym list  \citep{DBLP:journals/complexity/WangLZZC20} , heuristic rules \citep{DBLP:conf/inlg/KafleYK17}, encoder-decoder models \citep{DBLP:journals/corr/XiaQCBYL17} to reduce data imbalance by generating new samples. Recently there are methods that generate additional data samples using  Generative Adversarial Nets \citep{Goodfellow2014GenerativeAN,DBLP:conf/iclr/FedusGD18}, Majority-Minority translating  \citep{DBLP:conf/cvpr/KimJS20}, Easy data augmentation
techniques (EDA) \citep{DBLP:conf/emnlp/WeiZ19},  MixUp Augmentation \citep{DBLP:journals/corr/abs-1905-08941}, Contextual Augmentation \citep{fadaee-etal-2017-data, DBLP:journals/corr/abs-1805-06201} all of which describes forming new examples through some modification to existing examples. 

Alternately, there are also algorithm-level approaches. Primitive of these typically employ some form of cost variable and modify the classifier to output with reduced bias \citep{DBLP:journals/pai/Krawczyk16}. Then there works that change the loss functions to naturally account for data imbalance \citep{DBLP:journals/pami/LinGGHD20}. Typically, these loss functions manage the class imbalance problem by allotting more weights to complex or easily miss-classified examples.

In line with the above works, we focus on reducing imbalance in the context of hope speech detection. To this end, we contrast data level approaches like contrastive and back-translation-based augmentation with algorithmic strategies like Focal Loss while training the M-BERT model.

\subsection{Preprocessing Approaches} \label{s3}

While models and architectures play a crucial role in achieving high results, preprocessing tends to hold its unique position as it impacts extrinsic performance when integrated into the neural network architecture. There are a plethora of works beginning with \citep{Karlgren1041063} studying the impact of morphological analysis on creating word representations for synonymy detection,  \citep{Bullinaria2012ExtractingSR} present benefits of stemming and lemmatization before training word representations to show improvement in results. Similarly, \citep{DBLP:conf/kdweb/AngianiFFFIMM16} analyzes various preprocessing methods such as stopword removal, stemming, negation detection, emoticon preprocessing to find stemming to be most effective for the task of sentiment analysis. Similarly, \citep{DBLP:journals/corr/TraskML15} employ POS-disambiguated targets to show an improvement in the performance of a variety of tasks. In contrast, \citep{DBLP:conf/lrec/SaifFHA14} shows the negative impact on the classification performance of using the precompiled stoplist for the problem of sentiment analysis.

Recently there are works by \citep{DBLP:conf/emnlp/Ebert0S16} which introduce LAMB that show the advantage of word normalization on word similarity benchmarks.  \citep{DBLP:journals/access/ZhaoG17} observed that removing stopwords, numbers, and URLs can reduce noise but does not affect performance, whereas replacing negation and expanding acronyms can improve the classification accuracy. Alternatively, \citep{kuznetsov-gurevych-2018-text} examines the effect of lemmatization and POS typing on word embedding performance. \citep{DBLP:conf/wassa/PecarFSLB18} also highlights the importance of preprocessing when using user-generated content, with emoticons processing being the most effective. Finally, \citep{DBLP:conf/emnlp/Camacho-Collados18} shows an extensive evaluation on standard benchmarks of text categorization and sentiment analysis, highlighting significant variations in results across preprocessing techniques.

More recently, there are works \citep{DBLP:conf/acl/BabanejadAAP20} that comprehensively analyze the role of preprocessing techniques in affective analysis based on the word vector model. However, to the best of our knowledge, there are no works on preprocessing explicitly for language models and hope speech detection, which often takes in raw sentences as input. Accordingly, we introduce and study a simple overlapping word removal preprocessing algorithm that caters to the behavior of language models for the task of hope speech detection.

\section{Dataset and Experimental Setup}\label{experimental_setup}

\subsection{Dataset}\label{dataset}

In this work, we employ the Hope-speech dataset \citep{chakravarthi-2020-hopeedi} was released as a part of a shared task challenge on Language Technology for Equality, Diversity, and Inclusion \citep{chakravarthi-muralidaran-2021-findings}. The dataset is originally multilingual and consists of three subsets, namely English, Tamil, and Malayalam languages, with the latter two consisting of code-mixed text. In this work, we focus only on the English language. The English language subset consists of comments/posts from YouTube that offer support, reassurance, suggestions, inspiration, and insight divided into three classes, namely  \textit{Hope-speech}, \textit{Non-Hope-speech} or \textit{Not-English}. Overall dataset statistics are as shown in Table \ref{tab:Dataset}. As mentioned previously in Table \ref{tab:datadistrib}, the English dataset is highly imbalanced, with $91.28\%$ belonging to Non-Hope, in turn, exhibiting skew in the dataset. Further, considering that the proportion of samples in Not-English class is only 20, we remove Not-English class and focus on Hope-Speech and Non-Hope-Speech classes, respectively. In the rest of these documents, for convenience, the dataset will be referred to as HSDv2, and the classes will be referred to as Hope Class and Non-Hope Class, respectively.

\begin{table}[!htb]
\tbl{Statistics of Hope Speech Detection Dataset}
{ \begin{tabular}{@{}ccccc@{}}
\toprule
Language & Class & Training & Validation & Testing \\ \midrule
\multirow{3}{*}{English} & Hope & 1962 & 272 & 250 \\ \cmidrule{2-5} 
& Non-hope & 20778 & 2569 & 2593 \\ \cmidrule{2-5} 
& Not-English & 22 & 2 & 3 \\ 
\toprule
\end{tabular}}
\label{tab:Dataset}
\end{table}

% \begin{table}[!htb]
% \begin{center}
% \begin{minipage}{.7\textwidth}
% \caption{Statistics of Hope Speech Detection Dataset}
% \label{tab:Dataset}
% \begin{tabular}{@{}ccccc@{}}
% \toprule
% Language & Class & Training & Validation & Testing \\ \midrule
% \multirow{3}{*}{English} & Hope & 1962 & 272 & 250 \\ \cmidrule{2-5} 
% & Non-hope & 20778 & 2569 & 2593 \\ \cmidrule{2-5} 
% & Not-English & 22 & 2 & 3 \\ 
% \toprule
% \end{tabular}
% \end{minipage}
% \end{center}
% \end{table}

\subsection{Evaluation Metrics}

As mentioned previously in section \ref{intro}, currently in this work, we focus on to accurately classifying comments of Hope class. Accordingly, we use the following accuracy measures.

\begin{enumerate}

    \item \textbf{Precision}: is the proportion of correct classification to the number of incorrect classification.
    \begin{equation}
       Precision=\frac{TP}{TP+FP}
    \end{equation}
    \item \textbf{Recall}: is the proportion of correct classification to the number of missed entries.
    \begin{equation}
       Recall=\frac{TP}{TP+FN}
    \end{equation}
    \item \textbf{F1-Score}: is the harmonic mean of precision and recall.
    \begin{equation}
        F1-Score=2\times\frac{Precision\times Recall}{Precision+Recall}
    \end{equation}
\end{enumerate}

where True Positive (TP) corresponds to the number of comments correctly classified to Hope Class, True Negative (TN) accounts for the number of comments correctly classified to Non-Hope Class, False Positive (FN) considers the number of Hope-Speech comments wrongly classified to Non-Hope Class and False Negative (FN)  reflects the number of Non-Hope class comments incorrectly assigned to Hope class. 

Upon computing these measures for both Hope and Non\-Hope class, we calculate macro and weighed average using equations \ref{mcr} and \ref{wei}, where Score is replaced with either of precision, recall or F1-score and $W_{1}, W_{2}$ is calculated to the proportion of total sample belonging to each class.

\begin{equation}\label{mcr}
        Macro\ Average= 0.5 \times (Score_{Hope} + Score_{Non-Hope})
    \end{equation}
    
    \begin{equation}\label{wei}
        Weighed\ Average= W_{1} \times Score_{Hope} + W_{2} \times Score_{Non-Hope}
    \end{equation}

\section{Methodology}\label{tricks}

In this section, we present our baseline Multilingual-BERT model (section \ref{m1}), Focal Loss (section \ref{m2}), Contextual Data Augmentation (section \ref{m3}), Back-Translation Data Augmentation (section \ref{m3_1}) and Word Removal Pre-Processing Algorithm (section \ref{m4}).

\subsection{Model}\label{m1}
Multilingual-BERT (M-BERT) is widely used due to its ability to learn deep multilingual representation independent of overlap of vocabulary between train and test datasets \citep{DBLP:conf/acl/PiresSG19}. Accordingly, in this work, we employ M-BERT as our baseline.

Given, a sentence $S$ consisting of word sequence $\{w_{1},w_{2},....,w_{n}\}$ with a masked word $w_{i}$ ($i \in [1...n]$) in position $i$. Let $H^{0}$ be the input layer and $H^{j}$  represent intermediate layers defined as shown in Equations \ref{eqn:0-1} and \ref{eqn:0-2} respectively.

\begin{equation}\label{eqn:0-1}
H^{0} = [r_{1};...;r{i-1},r_{i},r{i+1};....r_{n}] + W_{p} \\
\end{equation}

\begin{equation}\label{eqn:0-2}
H^{j} = \textbf{SA}(H^{j-1}),\  j \in\ [1...K]
\end{equation}

where $r_{i}$ is embedding representation corresponding to word $w_{i}$, $W_{p} $ weights for the position embedding and \textbf{SA} corresponds to  Self-Attention \citep{DBLP:journals/corr/VaswaniSPUJGKP17} encoder that encodes input with $K$ Self-Attention layers. Originally, M-BERT is trained in form of Masked Language Model (MLM), where it predicts $w_{i}$ using Equation \ref{eqn:0-3}.

\begin{equation}\label{eqn:0-3}
P(w_{i}) = Softmax(Wh_{i}^{K})
\end{equation}

where $P(w_{i})$ denotes $P(w_{i}\|w_{1},....,w_{i-1},<MASK>,w_{i+1}, ...w_{n})$ and $W$ is the model parameter. For set if unlabelled text, $D=\{W^{i}\}$, M-BERT is originally, trained  on Wikipedia dump of 100 languages by maximizing 

\begin{equation}\label{eqn:0-4}
J = \sum_{i=1}^{N}\sum_{j=1}^{D}P[w_{i}^{j}]\\
\end{equation}

In this work, we finetune M-BERT for hope speech detection as follows.   Let $\textbf{h}_{i}$ $(i\in[1,...F])$ be the output of final hidden states of pretrained M-BERT for input hope speech word sequence $W$ and $\textbf{h}_{p}$ represent output from last hidden state. Let CE represent be the Cross-Entropy loss function that uses hope speech detection class labels $y \in [Hope,Non-Hope] $ to generate prediction probability $P(C_{i})$ over all the classes. Then classification is achieved by minimizing the following objective $J^{*}$ where $W_{h}$ corresponds to weight of fully connected layer. 

\begin{equation}\label{eqn:0-5}
P(C_{i}) = Softmax(W_{h}h_{p})
\end{equation}

\begin{equation}\label{eqn:0-6}
J^{*} = CE(y,P(C_{i}))
\end{equation}

\subsection{Focal Loss}\label{m2}
As discussed in section \ref{m1}, M-BERT is originally trained on Cross-Entropy loss. 
For a classification problem like Hope Speech Detection, Cross-Entropy (CE) is defined as shown in Equation \ref{eqn:1}.

\begin{equation}\label{eqn:1}
 CE(y,p) =
    \begin{cases}
      -log(p) & \text{if y=Hope}\\
      -log(1-p) & \text{otherwise}
    \end{cases}       
\end{equation}

In the above equation y corresponds ground-truth class and $p$ corresponds to probability class wise probability in range of $[0-1]$ indicating models estimate of class probability with label $y=Hope$.For simplicity let us define CE(p,y) = CE($p_{t}$), where  $p_{t}$ is as shown in Equation \ref{eqn:2}. 

\begin{equation}\label{eqn:2}
  p_{t} =
    \begin{cases}
      p & \text{if y=1}\\
      1-p & \text{otherwise}
    \end{cases}       
\end{equation}

While cross-entropy accounts for errors across the classes, it weighs predictions of all the classes equally, and it views weak ($p_{t}\approx0.5) $  and strong classifications ($p_{t}\geq0.7$)  equally, in turn assuming the error contribution to be similar. 

However, for the task of hope speech detection, CE loss is expected to be overwhelmed due to the usage of overlapping vocabulary in both Hope and Non-Hope classes. Moreover, wrong predictions of classes with minority samples are often ignored during training, resulting in limited generalization during actual testing. Additionally, Hope class usually consists of complex examples, which again would impact if trained with CE loss.

Focal loss addresses the imbalance issue and other side effects of class imbalance during training by using a modulating term to CE loss \citep{DBLP:journals/pami/LinGGHD20}. Such a modulating term helps CE focus on learning hard samples (samples belonging to minority classes and weakly classified). To this end, focal loss injects the approach of dynamic scaling policy, where the scaling factor ($\gamma$) decays to zero as confidence in the correct class increases for all the samples. Intuitively, $\gamma$ weighs down the contribution of easy examples during training and concentrates the model on hard examples. Formally, Focal Loss (FL) is defined as shown in \ref{eqn:3},  

\begin{equation}\label{eqn:3}
  FL(p_{t}) = -(1-p_{t})^{\gamma} log(p_{t}) 
\end{equation}

where $(1-p_{t})^{\gamma}$ is called modulating factor with $\gamma$ as a tunable parameter. From Equation \ref{eqn:3} we can see that a high value of $\gamma$ (ex: $\gamma >= 2$) weights down all samples and gradients that are approximately correct while converging on the examples which are yielding poor performance. On the other hand, having a too low $\gamma$ (ex: $\gamma$ = 0) reduces the impact of the weighting factor. 
However, the network will be incompetent to learn the problem for complex tasks like hope speech detection. Especially when the data is imbalanced, the network can easily be overwhelmed by the dominant data. In summary, $\gamma$ decides which training samples the network is most focused on at any given point during the training. As such, in this work, we restrict $\gamma$=$\{1,2\}$, thereby balancing learning of both poorly classified hate speech comments and easier ones, respectively. More details are presented in section \ref{results}.

\subsection{Contextual Data Augmentation}\label{m3}

In this work, we also study contextual word augmentation \citep{DBLP:conf/naacl/Kobayashi18}, where the words predicted by language models given the context surrounding the original sentence to be augmented. More specifically, we use the language model to calculate the word probability at a position $i$ based on its surrounding context. Let $S_{org}$ be the sentence of size $N$ with words $w_{i}=\{w_{1},....,w_{N}\}$ to be augmented, $LM$ be the language model, $K$ be hyperparameter to select number of candidate words to be selected for given word $w_{i}$ at position $i$ and $\{A_{min},A_{max}\}$ be minimum and maximum number of words to be augmented for $S_{org}$.  Then the contextual data augmentation procedure is as shown in Algorithm 1.

 \begin{algorithm}
    \caption{Contextual Data Augmentation}\label{reidd}
    \textbf{Input:} $S_{org}, LM, K, A_{min}, A_{max}$ \\
    \textbf{Output:} : Augmented Sentence $S_{aug}$ \\
    \textbf{Procedure:}
    \begin{itemize}
        \item[] Set $A_{cur}=0$
        \item[] \textbf{for} each word $w_{i}$ in $S_{org}$ 
        \item[]  \   \ Mask the word $w_{i}$ from position i
        \item[]  \   \ $WL$= Predict $K$ possible words for position $i$ using $LM$
        \item[] \    \ \textbf{if} $A_{cur} \in \{A_{min}, A_{max}\}$
        \item[] \   \    \ replace $w_{i}$ in $S$ by $WL[0]$
        \item[]  \   \   \ $S_{aug}= S_{aug} \cup S_{j}$
        \item[]  \  $\textbf{end for}$
        \item[] return $S_{aug}$
    \end{itemize}
\end{algorithm}

More specifically, in contextual data augmentation, we exploit Masked Language Modeling, which acts as the task of \textit{fill in the blank}, where we first mask a word and use a language model which exploits context words surrounding a mask to try to predict what the mask should be. To this end, for each position 'i', we mask the word and find new substitutes for word $w_{i}$ by sampling from a given probability distribution using a pre-trained language model. An example of contextual word augmentation is shown below with augmented words. In this study, we use the original BERT uncased model for augmentation and set  $A_{min}=3$ and $A_{max}=10$, respectively. 
\\
\\
\fbox{%
\centering
    \parbox{4.2in}{%
        \textbf{Original Comment:} Im only 11 but i love science and math i want to be a trauma surgeon specializing in nuero surgery. I definitely think and that all this will impower young girls like me or younger.\\

        \textbf{Contextually Augmented Comment:} I’m only 11 \textcolor{blue}{\textbf{months}} but \textcolor{blue}{\textbf{first}} i love science and math i want to be a \textcolor{blue}{\textbf{true}} trauma \textcolor{blue}{\textbf{rehabilitation}} surgeon specializing in neuro \textcolor{blue}{\textbf{stress}} surgery. I definitely \textcolor{blue}{\textbf{into}} think and that all this will impower young girls \textcolor{blue}{\textbf{around}} like me or younger.

    }
    }
\\

\subsection{Back Translation based Data Augmentation}\label{m3_1}
Previously we saw in section \ref{background_prelim} that few works use back-translation in the context of hope speech detection. To contrast with Focal Loss effectively, we also employ back translation based data augmentation in line with \citep{DBLP:conf/acl/SennrichHB16}. Back translation based augmentation focuses on using the fact that since LM's are encoder-decoder architectures, they already condition the probability distribution of the following target word on the previous target words. We can create additional synthetic data by automatically translating the original sentence to the target language and the target sentence back into the source language. This target language is called the intermediary/intermediate language.

Let $S= {S_{1},....S_{L}}$ be the set sentence of size $L$ with words $w_{i}=\{w_{1},....,w_{N}\}$, $LM=\{LM_{i},.....,LM_{K}\}$ be the K language models corresponding K different languages, then back translation based data augmentation is given as shown in Algorithm 2.

 \begin{algorithm}
    \caption{Back Translation based Data Augmentation}\label{reidd}
    \textbf{Input:} $S= {S_{1},....S_{L}}$, $LM=\{LM_{i},.....,LM_{K}\}$ \\
    \textbf{Output:} : Augmented Sentences $S_{aug}$ \\
    \textbf{Procedure:}
    \begin{itemize}
        \item[] $S_{aug}=\{\}$
        \item[] \textbf{for} each  $S_{i}$ in $S$
        \item[] \    \textbf{for} each  $LM_{i}$ in $LM$
        \item[] \   \    \ $S_{intermediate,i} = LM_{i}(S_{i})$
        \item[]  \   \   \ $S_{j} = LM_{j}(S_{intermediate,i})\  \forall \  \{i\leq k,j \leq k,i \neq j\}$
        \item[]  \   \   \ $S_{aug}= S_{aug} \cup S_{j}$
    \end{itemize}
\end{algorithm}
The example below shows back-translated augmentation with augmented words.
\\
\\
\fbox{%
\centering
    \parbox{4.2in}{%
       \textbf{Original Comment}: If you want to walk into a room and do not look like an engineer, you have to prove it what does that have to do with being a woman? If I walked in and \textcolor{blue}{\textbf{claimed to be a surgeon}} no one would believe me either. Most things have nothing to do with sexism...nnMany things would not have been possible without female engineers? Why \textcolor{blue}{\textbf{are we ok with}} that statement.\\
       
       \textbf{Back translated Comment:} If you want to walk into a room and not look like an engineer, you have to prove it, what does that have to do with being a woman? If I walked in and \textcolor{blue}{\textbf{said I'm a surgeon}}, no one would believe me either. Most things have nothing to do with sexism ... nn Many things would not have been possible without the engineers? Why \textcolor{blue}{\textbf{do we agree}} with that statement?
    }

    }
\\

\subsection{Word Removal Preprocessing Algorithm}\label{m4}
Multilingual-BERT shows multiple advantages, as mentioned in section \ref{m1}. Besides, it is reasonably stable against out-of-vocabulary words, mainly due to learning of semantic co-occurrence as part of its multi-head attention \citep{DBLP:journals/corr/abs-2012-15197}. However, in the case of hope speech detection, we find a significant overlap of vocabulary across hope-speech and non-hope-speech classes, respectively, i.e., specific tokens that are present for non-hope-speech are also present in hope-speech. More specifically, for English languages, the overlap statistics are as shown in Table \ref{tab:overlapstat}.

\begin{table}[!htb]
\tbl{Word Overlap Statistics for Hope Speech Detection Dataset}
{ \begin{tabular}{@{}cccc@{}}
\toprule
Data & Words in Hope & Words in Non-Hope & Overlapping Words \\ \toprule
Train & 7292 & 33458 & 5084 \\ \midrule
Validation & 2070 & 8645 & 1288 \\ \toprule
\end{tabular}}
\label{tab:overlapstat}
\end{table}

% \begin{table}[!htb]
% \centering
% \begin{minipage}{0.8\textwidth}
% \caption{Word Overlap Statistics for Hope Speech Detection Dataset}
% \label{tab:overlapstat}
% \begin{tabular}{@{}cccc@{}}
% \toprule
% Data & Words in Hope & Words in Non-Hope & Overlapping Words \\ \toprule
% Train & 7292 & 33458 & 5084 \\ \midrule
% Validation & 2070 & 8645 & 1288 \\ \toprule
% \end{tabular}
% \end{minipage}
% \end{table}

Such irregularity in word occurrences often impacts M-BERT's performance, as embeddings can be biased towards duplicated words' frequency. Furthermore, words that occur less frequently with Hope class, have less impact on the learned semantic meaning as they are sparsely distributed in the representation space \citep{DBLP:conf/emnlp/LiZHWYL20}. Besides such overlapping words, it leads to under-estimate model accuracy and ignores models' bias towards certain words. As such, we employ a word removal pre-processing algorithm.

Let $S_{c1}= \{S_{(1,c1)},....S_{(N,c1)}\}$ and $S_{c2}= \{S_{(1,c2)},....S_{(M,c2)}\}$ set of sentences from two subsets of data of sizes $N$ and $M$ respectively belonging to classes $c1$ and $c2$ respectively with each sentence consisting of words $w_{i}=\{w_{1},w_{2}....\}$. Then, we generate a inter class word-word 
occurrence matrix $WO$ $\in$ $R$ of size $L$x$Q$ using Equation \ref{eqn:3}. 

\begin{equation}\label{eqn:3}
WO_{i,j} = \frac{1}{L\times Q}\sum_{i=0}^{L}\sum_{j=0}^{Q}1\  \text{if}\ w_{i}=w_j
\end{equation}

where $L$ is the number of unique words in $S_{c1}$ and $Q$ is the number of the word in $S_{c2}$. To remove irrelevant words, remove words from such that  $WO_{i,j}\geq (\tau\times\frac{1}{L\times Q})$ words, and use the rest of the words. The summary of algorithmic steps is presented in Algorithm 3.

\begin{algorithm}
    \caption{Word Removal Preprocessing}\label{reidd}
    \textbf{Input:} $S_{c1}= \{S_{(1,c1)},....S_{(N,c1)}\}$, $S_{c2}= \{S_{(1,c2)},....S_{(M,c2)}\}$ , $\tau$ \\
    \textbf{Output:} : $S_{(c1,cleaned)},S_{(c2,cleaned)}$ \\
    \textbf{Procedure:}
    \begin{itemize}
        \item[--] Extract $L,Q$ unique words from $S_{c1},S_{c2}$ respectively.  
        \item[--] Compute $WO_{i,j} = \frac{1}{L\times Q}\sum_{i=0}^{L}\sum_{j=0}^{Q}1\  \text{iff}\ w_{i}=w_j$
        \item[--] $WO_{i,j} = \begin{cases}
           WO_{i,j}\ \text{if}\ WO_{i,j}\geq (\tau\times\frac{1}{L\times Q}) \\
           0\ \text{otherwise}
        \end{cases}
        $
        \item[--] Remove words  $w_{i},w_j$ if  $WO_{i,j}=0$ from $S_{c1},S_{c2}$
        \item[--] $S_{(c1,cleaned)},S_{(c2,cleaned)}$ = $S_{c1},S_{c2}$
    \end{itemize}
\end{algorithm}

\section{Experiments and Results}\label{results}

In this section, we start by explaining the baseline experiments on M-BERT, followed by improvements and characteristics of each of the strategies proposed in section \ref{tricks}.

\subsection{Baseline}\label{base_res}

We start with the development of the baseline model of M-BERT, for which we fine-tune using the HSDv2 from section \ref{dataset}. To this end, we train for ten epochs using cross-entropy loss with 1k warm\-up steps, batch size of 8, the learning rate of $3.0\times10^-5$, the sequence length of 160, and ADAM \citep{DBLP:journals/corr/KingmaB14} optimizer with clip threshold of 1.0 and epsilon value of $1\times10^{-8}$. Detailed results on both validation and test splits are as shown in Table \ref{tab:baseline} with respective Precision, Recall, and F1-Score with Macro Average results. Analysis of results shows various interesting details leading towards the requirement of methods in section \ref{tricks}. 

\begin{table}[!htb]
\tbl{Baseline Results of M-BERT for English Hope Speech Detection}
{ \begin{tabular}{cccccc} 
\toprule
Language & Data & Class & Precision & Recall & F1-Score \\ 
\toprule
\multirow{8}{*}{English}  &\multirow{4}{*}{Validation Data} & Hope (271)& 0.6512 & 0.6020 & 0.6257 \\ %\cmidrule{3-6}
& & Non-hope(2569) &0.7002 &0.7425 &0.7207 \\ %\cmidrule{3-6}
& & Macro Average(2840) & 0.6757 & 0.6722 & 0.6732 \\ %\cmidrule{3-6}
& & Weighted Average & 0.9275&0.9306 &0.9289 \\ \cmidrule{2-6}
& \multirow{4}{*}{Test Data} & 
Hope (250)& 0.6658  & 0.6088 & 0.6360 \\ %\cmidrule{3-6}
& & Non-hope(2593) &0.6762 &0.7278 &0.7011 \\ %\cmidrule{3-6}
& & Macro Average(2843) & 0.6710 & 0.6683 & 0.6686 \\ %\cmidrule{3-6}
& & Weighted Average & 0.9277&0.9307 &0.9290 \\ \toprule
\end{tabular}}
\label{tab:baseline}
\end{table}

% \begin{table}[!htb]
% \centering
% \begin{minipage}{320pt}
% \caption{Baseline Results of M-BERT for English Hope Speech Detection }
% \label{tab:baseline}
% \begin{tabular}{cccccc} 
% \toprule
% Language &  & Class & Precision & Recall & F1-Score \\ 
% \toprule
% \multirow{2}{*}{English}  & Validation Data & \begin{tabular}[c]{@{}c@{}}Hope (271)\\ Non-hope (2569)\\ Macro Average (2840)\\Weighted Average \end{tabular} & \begin{tabular}[c]{@{}c@{}}0.6512\\ 0.7002\\ 0.6757\\ 0.9275\end{tabular} & \begin{tabular}[c]{@{}c@{}}0.6020\\ 0.7425\\ 0.6722\\0.9306\end{tabular} & \begin{tabular}[c]{@{}c@{}}0.6257\\ 0.7207\\ 0.6732\\0.9289
% \end{tabular} \\ \cline{2-6} 
%  & Test Data & \begin{tabular}[c]{@{}c@{}}Hope (250)\\ Non-Hope (2593)\\ Macro Average (2843)\\Weighted Average\end{tabular} & \begin{tabular}[c]{@{}c@{}}0.6658\\ 0.6762\\ 0.6710\\0.9277\end{tabular} & \begin{tabular}[c]{@{}c@{}}0.6088\\ 0.7278\\ 0.6683\\0.9307\end{tabular} & \begin{tabular}[c]{@{}c@{}}0.6360\\0.7011\\0.6686\\0.9290\end{tabular} \\ 
% \toprule
% \end{tabular}
% \end{minipage}
% \end{table}

\begin{figure}[!htb]
    \centering
    \scalebox{0.7}{
    \includegraphics{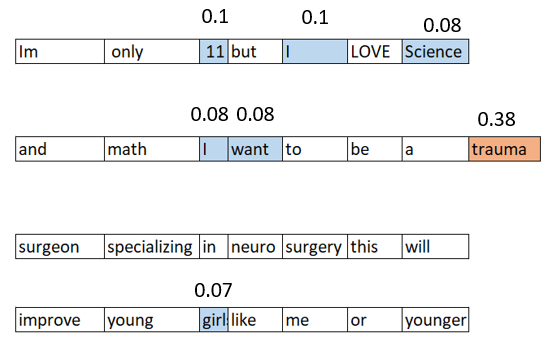}}
    \caption{LIME Predictions for hard example.  }
    \label{fig:1}
\end{figure}

\begin{figure}[!htb]
    \centering

    \scalebox{0.7}{
    \includegraphics{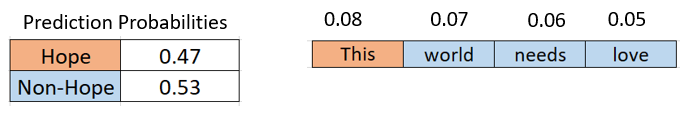}}
    \caption{LIME predictions showcasing word bias.}
    \label{fig:2}
\end{figure}

\begin{itemize}
    \item Firstly, we can see that the overall results are quite low, with 0.6732 and 0.6686 for both validation and test sets.
    \item Furthermore, we can see that between the classes of Hope and Non-Hope, the net results differ by a large margin ($\geq 0.10$ and $\geq 0.07$\% in F1-Score). To this end, we can attribute such a large difference to imbalanced samples for the Hope class.
    \item At the same time, despite the large difference in F1, we can argue that M-BERT being a strong baseline as it shows balanced F1, despite having 100$\times$ fewer samples for Hope class.
    \item We further explicitly examine results on each class individually using input perturbation-based explanation method involving Local Interpretable Model Agnostic Explanations (LIME) \citep{lime}. Examples are as shown in Figures  \ref{fig:1} and  \ref{fig:2} respectively. From a manual analysis of such an explanation, we find that many of the problems that we defined earlier in section \ref{intro} are valid. More specifically, we find that. 
    \begin{itemize}
        \item[--]    Sheer imbalance on dataset often flips the hard samples (See Figure \ref{fig:1}), which suggests incorporating strategies for handling imbalance as part of the overall training pipeline.  %Add surgeon example
        \item[--]  Besides, we can also see some words impacting the final predictions, leading to erroneous results. For example, the word \textit{this} in Figure  \ref{fig:2} is shown to contribute to the final results, despite having no semantically meaningful information. Such results acknowledge the need to revisit and add preprocessing stages in hope speech detection. 
    \end{itemize}
\end{itemize}

To summarize from our baseline, we find
\begin{itemize}
    \item Overall, F1-Scores differ by a large margin for Hope and Non-Hope classes.
    \item M-BERT despite having 100$\times$ lesser sample for Hope class, shows descent performance, suggesting its suitability as a strong baseline for Hope Speech Detection.
    \item Our initial postulation of issues related to data imbalance and word overlap is true and often results in flipping predictions towards the wrong class.
\end{itemize}

\subsection{Effect of Focal Loss}\label{focal_res}

Having examined the baseline using M-BERT in section \ref{base_res} where we touched upon the issues due to imbalance via LIME interpretations, we now start by focusing on establishing an approach to reduce the impact of data imbalance. While data imbalance may explain low F1 with M-BERT, to establish this quantitatively, we start by replacing cross-entropy with focal loss in M-BERT's training pipeline. We begin by examining Precision, Recall, and F1-Score with both Macro and Weighed Average measures as shown in Table \ref{tab:flenglish} with $\gamma=2$.

To start with, we can see that compared to baseline with focal loss, the F1-Macro has improved to 0.7860 and 0.7859 on validation and test sets, respectively. This is approximately 0.12 improvement over baseline. The following shows examples of various comments which initially were wrongly identified as Non-Hope by cross-entropy loss but were corrected when coupled with focal loss.
\\
\\
\fbox{%
\centering
    \parbox{4.2in}{%
       \textbf{Example 1}: This was so wonderful!\\
       \textbf{Example 2}: I watched this to become educated on this matter\\
       \textbf{Example 3}: This made me cry. Thank you for posting it! This gives me hopes for the future.
    }
    % \caption{Example of Contextual Data Augmented Comments.}  
    }
\\

\begin{table}[!htb]
\tbl{Results of M-BERT with Focal Loss HSDv2 Dataset}
{\begin{tabular}{cccccc}
\toprule
Model & Data & Class & Precision & Recall & F1-Score \\ \toprule

\multirow{8}{*}{\makecell{M-BERT with \\ Focal Loss \\(Gamma=1)\\(binary class)}}
&\multirow{4}{*}{Validation Data} &Hope&0.6125&0.6125&0.6125 \\ %\cmidrule{3-6}
& &Non-hope & 0.9591 & 0.9591 & 0.9591 \\ %\cmidrule{3-6}
& &Macro average &0.7858 &0.7858 &0.7858 \\ %\cmidrule{3-6}
& &Weighted average & 0.9261&0.9261 &0.9261\\ \cmidrule{2-6}
&\multirow{4}{*}{Test Data} &Hope&0.5809&0.5600&0.5703 \\ %\cmidrule{3-6}
& &Non-hope & 0.9577& 0.9610 & 0.9594 \\ %\cmidrule{3-6}
& &Macro average &0.7693 &0.7605 &0.7648 \\ %\cmidrule{3-6}
& &Weighted average & 0.9246&0.9258 &0.9252\\ \midrule

\multirow{8}{*}{\makecell{M-BERT with \\ Focal Loss \\(Gamma=2)\\(binary class)}}
&\multirow{4}{*}{Validation Data} &Hope&0.6423&0.5830&0.6112 \\ %\cmidrule{3-6}
& &Non-hope & 0.9561 & 0.9657 & 0.9609 \\ %\cmidrule{3-6}
& &Macro average &0.7992 &0.7744 &0.7860 \\ %\cmidrule{3-6}
& &Weighted average & 0.9262&0.9292 &0.9275\\ \cmidrule{2-6}
&\multirow{4}{*}{Test Data} &Hope&0.6052&0.5640&0.5839 \\ %\cmidrule{3-6}
& &Non-hope & 0.9571& 0.9645 & 0.9608 \\ %\cmidrule{3-6}
& &Macro average &0.8031 &0.7710 &0.7859 \\ %\cmidrule{3-6}
& &Weighted average & 0.9662&0.9292 &0.9277\\ \toprule

\end{tabular} 
}
\label{tab:flenglish}
\end{table}

The above examples are not straightforward when it comes to identification as Hope Speech. This is because some of the words \textit{so, educated, cry} etc., which are vital in deciding context, are prone to non-uniformities which can skew the results.  This moderately supports our hypothesis that penalizing complex examples is critical in hope speech detection to sufficiently understand the context when data is imbalanced.

Although the results are significantly higher, we can see much of the improvement in results are concerning Non-Hope class, where for Non-Hope class, the results increase by approximately 0.22 with the drop in results for the Hope class by 0.01. We believe this because, despite handling imbalance, the $\gamma$ factor plays a crucial lesson in generalization.  

More specifically, we can see that with $\gamma=2$, there is an additional improvement by addition 0.01 for the Hope class, which begs to question if this should be increased further. In this regard, we believe that a higher $\gamma$ may focus on complex examples. Fixing this value to a static number is not helpful as this may lead to convergence issues as mentioned in section \ref{m2}. 

Moreover, with the low gamma value, the network will not learn any complex examples at all. Thus, at the beginning of this study, we examined to strike a balance by studying both $\gamma=\{1,2\}$. However, we can see that much of focal loss tried to improve overall F1 by focusing on complex examples of Non-Hope Speech.  Following are some of the samples from the Hope class that were miss-classified as Non-Hope.
\\
\\
\fbox{%
\centering
    \parbox{4.2in}{%
       \textbf{Example 1}: Im only 11 but I LOVE science and math I want to be a trauma surgeon specializing in nuero surgery. I definitely think that this will empower young girls like me or younger.\\
       \textbf{Example 2}: its not common for women to work as blue-collar job in construction or plumbing.. Most of 'em are in the professional\\
       \textbf{Example 3}: why can someone not agree with something? She is not hurting anyone with her opinion as long as she is not harassing people.
    }
    % \caption{Example of Contextual Data Augmented Comments.}  
    }
\\

Comparing the miss classified ones to the comments that were corrected, we can see one exact problem; namely, sentences of larger length, predominantly filled with words that are often part of Non-Hope class such as trauma, plumbing, construction, hurting, etc. In our work, upon analysis, we find that more than 20\% of total errors introduced by focal loss for hope speech class have these characteristics. We believe such an issue is because of fixed $\gamma$, whereas when words that frequent with Non-Hope class appear with Hope class, the classification gets penalized. Such repetitive penalization for Hope class texts due to Non-hope class leads to delay in convergence, causing performance drop. To summarize, our findings are 

\begin{itemize}
    \item Focal Loss with $\gamma=1$ has an overall result of 0.7858, and 0.7648 and $\gamma=2$ had 0.7860 and 0.7859 across validation and test sets, respectively, indeed highlighting the complexity of the Hope class.
    \item Further, we can see that while focal loss increases the overall F1, we can see that the Hope class results are lower than the baseline. To this end, we believe this may be attributed to fixed constant penalization upon miss-classification due to words frequenting with Non-Hope classes.
    \item Also, we find, despite lower than initial results, Focal loss still helps in correcting errors in  Hope class.
\end{itemize} 

\subsection{Effect of Data Augmentation}\label{Aug_res}
While focal loss tackles data imbalance in the context of training, data augmentation reduces imbalance by generating more data. Accordingly, as mentioned \ref{m3} and \ref{m3_1}, we employ two data augmentation strategies by coupling them with both cross-entropy loss and focal loss. Though some existing works in hope speech detection aim to increase performance through using back translation, the results obtained are fairly limited. So, for back translation, we also include two languages, Spanish and French, as an intermediary, owing to their word order being similar to English. While back translation is a fairly straightforward idea, we believe that comments from social media often suffer from the micro-domain issue, where each has its style, impacting overall back translation. In particular, one or more back translations may also lead to a drop in results. The dataset statistics post each of the augmentation strategies is as shown in Table \ref{tab:aug_data}.

\begin{table}[!htb]
\tbl{Statistics of HSDv2 after Data Augmentation}
{ \begin{tabular}{cccc}
\toprule
\multirow{2}{*}{\textbf{}} & \multicolumn{3}{c}{\textbf{Train}} \\ \cmidrule{2-4} 
 & \textbf{Original} & \textbf{Contextual Augmented} & \textbf{Back-translation Augmented} \\ \midrule
\textbf{Train} & 1962 & 3924 & 5886 \\ \toprule
\end{tabular}}
\label{tab:aug_data}
\end{table}
% \begin{table}[!htb]
% \centering
% \caption{Statistics of HSDv2 after Data Augmentation}
% \label{tab:aug_data}
% \begin{tabular}{cccc}
% \toprule
% \multirow{2}{*}{\textbf{}} & \multicolumn{3}{c}{\textbf{Train}} \\ \cmidrule{2-4} 
%  & \textbf{Original} & \textbf{Contextual Augmented} & \textbf{Back-translation Augmented} \\ \midrule
% \textbf{Train} & 1962 & 3924 & 5886 \\ \toprule
% \end{tabular}
% \end{table}

To begin with, we experiment using M-BERT with cross-entropy loss function and employ contextual word embedding augmentation. The results are as shown in Table \ref{tab:augenglish}. Comparing these results with a baseline from Table \ref{tab:baseline}, we can see that contextual augmentation improves the overall F1-Macro by 0.1. However, we can see that net improvement post augmentation is seen for only the Non-Hope class. Meanwhile, when trained with Focal loss, the net results are a bit on the lower side compared to that of cross-entropy loss (See table \ref{tab:flenglish}), yet with the overall improvement of 0.07 F1 again significant change observable only for Non-Hope class.

% \begin{figure}[!htb]
%   \centering
%   \subfloat{\includegraphics[width=.4\textwidth]{figures/val_aug1.png}}\quad
%   \subfloat{\includegraphics[width=.4\textwidth]{figures/test_aug1.png}}\\
%   \subfloat{\includegraphics[width=.4\textwidth]{figures/val_aug1_focal.png}}\quad
%   \subfloat{\includegraphics[width=.4\textwidth]{figures/test_aug1_focal.png}}
%   \caption{Confusion Matrix for M-BERT with Contextual Word Embedding.}
%   \label{fig:sub1x}
% \end{figure}

While the reason for the drop in Hope class performance is in line with that of discussion from section \ref{focal_res}, we believe that much of augmentation introduces words or phrases that are often associated with Non-Hope class. To this end, we analyze the statistics of augmentation and some of the words used to find that upon augmentation. We introduce 2$\times$ new sentences out of which close to 70\% augmented words can be associated with non-hope than hope. For example, after using contextual word augmentation in the below statement, the predictions of Hope class were mispredicted with both M-BERT and M-BERT with focal loss. 
\\
\\
\fbox{%
\centering
    \parbox{4.2in}{%
       \textbf{Original Comment}:  It's pretty simple. A large portion of the world is anti-gay and the parades are a way to show that gay people aren't going anywhere and demand the same rights that everyone else has. It's especially useful for the gay teenagers that are put out onto the street by their christian families and feel like they dont belong anywhere. Being able to see that people like you are celebrating and proud of who they are when everyone around you wants to make you feel ashamed for being gay is something that saves lives. Hope that answer explains part of it \\
       \textbf{Contextual Word Augmented Comment:} this it's pretty simple. a large portion of the world is anti - gay and the parades are a way to mock show that no gay people aren't going anywhere and demand the same rights that everyone else has. it's especially useful for the gay teenagers that are now put over out onto the street by their private christian families and feel like they dont belong anywhere. being able to see that people like you are celebrating and proud of who they are when everyone passing around you wants to make you feel ashamed for being secretly gay is something online that saves lives. hope that answer explains part of it
    }
    % \caption{Example of Contextual Data Augmented Comments.}  
    }
\\

\begin{table}[!htb]
\tbl{Results of M-BERT with HSDv2 with various Data Augmentation approaches}
{ \begin{tabular}{cccccc}
\toprule

Model & Data & Class & Precision & Recall & F1-Score \\ \toprule

\multirow{8}{*}{\makecell{M-BERT with\\Contextual embedding}}
&\multirow{4}{*}{Validation Data} 
& Hope&0.6667&0.5092&0.5774 \\ %\cmidrule{3-6}
& &Non-hope & 0.9495 & 0.9731 & 0.9612 \\ %\cmidrule{3-6}
& &Macro average &0.8081 &0.7412 &0.7693 \\ %\cmidrule{3-6}
& &Weighted average & 0.9225&0.9289 &0.9245\\ \cmidrule{2-6}
&\multirow{4}{*}{Test Data} &Hope&0.6308&0.5400&0.5819 \\ %\cmidrule{3-6}
& &Non-hope & 0.9563& 0.9695 & 0.9628 \\ %\cmidrule{3-6}
& &Macro average &0.7935 &0.7548 &0.7724 \\ %\cmidrule{3-6}
& &Weighted average & 0.9276&0.9318 &0.9294\\ \midrule

\multirow{8}{*}{\makecell{M-BERT with\\Back translation\\ (French)}}
&\multirow{4}{*}{Validation Data} &Hope&0.6314&0.5941&0.6122 \\ %\cmidrule{3-6}
& &Non-hope & 0.9574 & 0.9634 & 0.9604 \\ %\cmidrule{3-6}
& &Macro average &0.7944 &0.7788 &0.7863 \\ %\cmidrule{3-6}
& &Weighted average & 0.9263&0.9282 &0.9272\\ \cmidrule{2-6}
&\multirow{4}{*}{Test Data} &Hope&0.6188&0.5520&0.5835 \\ %\cmidrule{3-6}
& &Non-hope & 0.9573& 0.9672 & 0.9622 \\ %\cmidrule{3-6}
& &Macro average &0.7880 &0.7596 &0.7729 \\ %\cmidrule{3-6}
& &Weighted average & 0.9275&0.9307 &0.9289\\ \midrule

\multirow{8}{*}{\makecell{M-BERT with\\Contextual embedding \\and Back translation}}
&\multirow{4}{*}{Validation Data} 
& Hope&0.6429&0.5314&0.5818 \\ %\cmidrule{3-6}
& &Non-hope & 0.9515 & 0.9689 & 0.9601 \\ %\cmidrule{3-6}
& &Macro average &0.7972 &0.7501 &0.7709 \\ %\cmidrule{3-6}
& &Weighted average & 0.9220&0.9271 &0.9240\\ \cmidrule{2-6}
&\multirow{4}{*}{Test Data} &Hope&0.5822&0.4960&0.5356 \\ %\cmidrule{3-6}
& &Non-hope & 0.9521& 0.9657 & 0.9588 \\ %\cmidrule{3-6}
& &Macro average &0.7671 &0.7308 &0.7472 \\ %\cmidrule{3-6}
& &Weighted average & 0.9196&0.9244 &0.9216\\ \midrule

\multirow{8}{*}{\makecell{M-BERT with\\Back translation\\ (Spanish)}}
&\multirow{4}{*}{Validation Data} &Hope&0.6652&0.5572&0.6064 \\ %\cmidrule{3-6}
& &Non-hope & 0.9541 & 0.9704 & 0.9622 \\ %\cmidrule{3-6}
& &Macro average &0.8096 &0.7638 &0.7843 \\ %\cmidrule{3-6}
& &Weighted average & 0.9265&0.9310 &0.9282\\ \cmidrule{2-6}
&\multirow{4}{*}{Test Data} &Hope&0.6308&0.5400&0.5819 \\ %\cmidrule{3-6}
& &Non-hope & 0.9563& 0.9695 & 0.9628 \\ %\cmidrule{3-6}
& &Macro average &0.7935 &0.7548 &0.7724 \\ %\cmidrule{3-6}
& &Weighted average & 0.9276&0.9318 &0.9294\\
\toprule
\end{tabular}}
\label{tab:augenglish}
\end{table}

Back translations are often beneficial as intermediary languages may introduce new words and, in fact, wider contexts. Corresponding experiments and results are as shown in Table \ref{tab:augenglish} and \ref{tab:fl&augenglish} with both French and Spanish as intermediate languages. Comparing these results against \ref{tab:baseline} and \ref{tab:flenglish}, we can see the overall results are significantly higher. Specifically, we notice that back translation introduces an improvement of 0.11 over M-BERT. However, compared to M-BERT with focal loss, we find that net improvement to be limited. Besides, we can see that in line with contextual word augmentation, the results are improved for Non-Hope class and dropped for the Hope class. Such an issue is similar to the one mentioned earlier, with examples below showing such an impact in prediction. Especially, we could see that in the dataset, the words such as \textit{practically, believing, our}  flipped predictions towards Non-Hope class.
\\
\\
\fbox{%
\centering
    \parbox{4.2in}{%
       \textbf{Original Comment}:  I'm a Buddhist...!    ALL LIVES MATTER...!	 \\
       \textbf{Back-translation Augmented:}  I'm \textcolor{blue}{\textbf{practically}} a \textcolor{blue}{\textbf{believing}} buddhist...! our all lives forever matter...!
    }
    % \caption{Example of Contextual Data Augmented Comments.}  
    }
\\

Also, comparing across different languages used, we find that French is more
useful for back-translation-based augmentation than Spanish. This we believe is
because French word tends to be identified by their surface forms similar to English.
Specifically, we can see that using French as back translation tends to improve overall
results by 0.03 and 0.01 respectively on M-BERT and M-BERT with focal loss. Finally,
the net performance when all of them are put together, i.e., both augmentations used
together the results are intermediate between the respective baselines, and the results of back translation with french as an intermediate language.

\begin{table}[!htb]
\tbl{Results of M-BERT with focal loss various Data Augmentation approaches on HSDv2.}
{ \begin{tabular}{cccccc}
\toprule

Model & Data & Class & Precision & Recall & F1-Score \\ \toprule

\multirow{8}{*}{\makecell{M-BERT with\\Word embedding\\(Focal Loss)}}
&\multirow{4}{*}{Validation Data} 
& Hope&0.6667&0.5092&0.5774 \\% \cmidrule{3-6}
& &Non-hope & 0.9495 & 0.9731 & 0.9612 \\ %\cmidrule{3-6}
& &Macro average &0.8081 &0.7412 &0.7693 \\ %\cmidrule{3-6}
& &Weighted average & 0.9225&0.9289 &0.9245\\ \cmidrule{2-6}
&\multirow{4}{*}{Test Data} &Hope&0.6105 &0.4640 &0.5273 \\ %\cmidrule{3-6}
& &Non-hope & 0.9495 &0.9715 &0.9604 \\ %\cmidrule{3-6}
& &Macro average &0.7800 &0.7177 &0.7438 \\ %\cmidrule{3-6}
& &Weighted average & 0.9197 &0.9268 &0.9223\\ \midrule

\multirow{8}{*}{\makecell{M-BERT with\\Word embedding \\and Back translation \\(Focal Loss)}}
&\multirow{4}{*}{Validation Data} &Hope&0.6329 &0.5535 &0.5906 \\ %\cmidrule{3-6}
& &Non-hope & 0.9535 &0.9661 &0.9598\\ %\cmidrule{3-6}
& &Macro average &0.7932& 0.7598 &0.7752 \\ %\cmidrule{3-6}
& &Weighted average & 0.9229 &0.9268 &0.9246\\ \cmidrule{2-6}
&\multirow{4}{*}{Test Data} &Hope&0.5965 &0.5440 &0.5690 \\ %\cmidrule{3-6}
& &Non-hope & 0.9564 &0.9645 &0.9604 \\ %\cmidrule{3-6}
& &Macro average &0.7764 &0.7543 &0.7647 \\ %\cmidrule{3-6}
& &Weighted average & 0.9248 &0.9275 &0.9260\\ \midrule

\multirow{8}{*}{\makecell{M-BERT with\\Back translation\\ (Spanish) \\(Focal Loss)}}
&\multirow{4}{*}{Validation Data} 
& Hope& 0.6652 & 0.5572 &0.6064\\ %\cmidrule{3-6}
& &Non-hope &0.9541 &0.9704 &0.9622  \\ %\cmidrule{3-6}
& &Macro average & 0.8096 &0.7638 &0.7843\\ %\cmidrule{3-6}
& &Weighted average & 0.9265& 0.9310& 0.9282\\ \cmidrule{2-6}
&\multirow{4}{*}{Test Data} &Hope&0.6308 & 0.5400 & 0.5819 \\ %\cmidrule{3-6}
& &Non-hope &0.9563 & 0.9695 & 0.9628  \\ %\cmidrule{3-6}
& &Macro average &0.7935 & 0.7548 & 0.7724 \\ %\cmidrule{3-6}
& &Weighted average &0.9276 & 0.9318 & 0.9294 \\ \midrule

\multirow{8}{*}{\makecell{M-BERT with\\Back translation\\ (French)\\(Focal Loss)}}
&\multirow{4}{*}{Validation Data} &Hope& 0.6667 & 0.5830 & 0.6220\\ %\cmidrule{3-6}
& &Non-hope &0.9566 & 0.9692 & 0.9629  \\ %\cmidrule{3-6}
& &Macro average & 0.8116 & 0.7761 & 0.7925\\ %\cmidrule{3-6}
& &Weighted average & 0.9289 & 0.9324 & 0.9304\\ \cmidrule{2-6}
&\multirow{4}{*}{Test Data} &Hope&0.6650 & 0.5480 & 0.6009 \\ %\cmidrule{3-6}
& &Non-hope & 0.9571 & 0.9734 & 0.9652 \\ %\cmidrule{3-6}
& &Macro average &0.8111 & 0.7607 & 0.7830 \\ %\cmidrule{3-6}
& &Weighted average &0.9315 & 0.9360 & 0.9332 \\
\toprule
\end{tabular}}
\label{tab:fl&augenglish}
\end{table}

To summarize, we find the following
\begin{itemize}
    \item Both contextual and back-translation based word augmentation improves the F1-Macro by 10\% and 11\% respectively, yet they both show the issue of improving only the Non-Hope class. 
    \item Also, both augmentation show errors due to introducing words that are commonly associated with Non-Hope class.
    \item Focal loss, when coupled with Augmented data, has a substantial impact on the results.
    \item Among Spanish and French the latter seems more beneficial for back-translation based augmentation with 3\% more improvement in F1-Macro. 
    \item Finally, with both strategies together, the results are higher than baseline M-BERT and M-BERT with focal loss but lower than back-translation.
\end{itemize}

\subsection{Effect of Word Removal}\label{removal_res}

As seen in section \ref{base_res}, word overlap often biases the results towards the majority class, especially in the case of significant class imbalance. Previously in section \ref{m4}, we saw that between the classes of Hope and Non-Hope, 70\% of words present in Hope class is present in Non-Hope class.

We hypothesize that overlapping word removal across classes should not hurt model generalization and is expected to have no worse result compared to that of the baseline model in section \ref{m1}. As such, we start by executing a word removal algorithm with $\tau=50$. The results so obtained are as shown in Table \ref{tab:wr} using M-BERT trained on both cross-entropy loss and focus loss, respectively.

\begin{table}[!htb]
\tbl{Results of M-BERT with overlapping word removal.}
{ \begin{tabular}{@{}cccccc@{}} 
\toprule
Loss function & Data & Class & Precision & Recall & F1-Score \\ 
\toprule
\multirow{8}{*}{Cross Entropy}
&\multirow{4}{*}{Validation Data} 
& Hope(271)&0.9449&0.9876&0.9663 \\ %\cmidrule{3-6}
& &Non-hope(2569) & 0.8824 & 0.9946 & 0.9385 \\ %\cmidrule{3-6}
& &Macro average(2840) &0.9125 &0.9911 &0.9518 \\%\cmidrule{3-6}
& &Weighted average(2840) & 0.9835&0.9838 &0.9836\\ \cmidrule{2-6}
&\multirow{4}{*}{Test Data} 
&Hope(250)&0.9362&0.9885&0.9623 \\ %\cmidrule{3-6}
& &Non-hope(2593) & 0.8800& 0.9942 & 0.9371 \\ %\cmidrule{3-6}
& &Macro average(2843) &0.9072 &0.9913 &0.9493 \\ %\cmidrule{3-6}
& &Weighted average(2843) & 0.9839&0.9842 &0.9839\\ \midrule

\multirow{8}{*}{Focal Loss}
&\multirow{4}{*}{Validation Data} 
&Hope(271)&0.9555&0.9861&0.9708 \\ %\cmidrule{3-6}
& &Non-hope(2569) & 0.8676 & 0.9957 & 0.9317 \\ %\cmidrule{3-6}
& &Macro average(2840) &0.9094 &0.9909 &0.9502 \\ %\cmidrule{3-6}
& &Weighted average(2840) & 0.9832&0.9835 &0.9831\\ \cmidrule{2-6}
&\multirow{4}{*}{Test Data} 
&Hope(250)&0.9442&0.9885&0.9664 \\ %\cmidrule{3-6}
& &Non-hope(2593) & 0.8800& 0.9950 & 0.9375 \\% \cmidrule{3-6}
& &Macro average(2843) &0.9110 &0.9917 &0.9514 \\% \cmidrule{3-6}
& &Weighted average(2843) & 0.9846&0.9849 &0.9846\\
\toprule
\end{tabular}}
\label{tab:wr}
\end{table}

As can be seen from Table \ref{tab:wr}, with the removal of overlapping words, the overall results tend to improve significantly where the test corpus with a low percentage of seen words tend to have a high F1-Macro. This suggests the system can recognize necessary context and efficiently classify the contents across different classes. 

Also, comparing these results from Table \ref{tab:baseline}, the F1-Macro with seen words on both classes tend to lower results significantly with cross-entropy loss and suggest some form of memorization and poor generalization. In particular, we can see the Precision, Recall and F1-Score to have almost 0.3 higher results for the Hope class. Of the two classes, we can see that the Hope class improves by an additional margin of 10\% over Non-hope. As a result, both cross the 0.9  in F1. This may be because overlapping words across classes has the effect of biasing the pre-trained language models towards specific types of examples. This results in lower generalization with overlapping words.

\begin{table}[!htb]
\tbl{Example Predictions after using Overlapping Word Removal Algorithm with M-BERT coupled with Cross-Entropy and Focal Loss}
{\begin{tabular}{ccccc}
\toprule
Original Tweet & Modified Tweet & Label & \makecell{Word Removal \\ Prediction \\CE Loss} & \makecell{Word Removal \\ Prediction \\Focal Loss} \\ \toprule
\makecell{“God gave us a choice \\ and my choice is to love”\\ nnPROTECT THIS CHILD} & \makecell{“God love”nn\\ PROTECT THIS CHILD} & Hope & Non\_Hope& Hope \\ \midrule
\makecell{I love how kids can react \\so positively to this} & kids react positively & Hope & Non\_Hope & Hope \\ \midrule
\makecell{Good thing my country \\ never had slavery} & \makecell{Good thing country \\ slavery} & Hope & Hope & Non\_Hope\\ \toprule
\end{tabular}}
\label{tab:wordrem}
\end{table}

% \begin{table}[!htb]
% \begin{minipage}{\textwidth}
% \caption{Example Predictions after using Overlapping Word Removal Algorithm with M-BERT coupled with Cross-Entropy and Focal Loss}
% \label{tab:wordrem}
% \resizebox{\textwidth}{!}{%
% \begin{tabular}{@{}ccccc@{}}
% \toprule
% Original Tweet & Modified Tweet & Label & \begin{tabular}[c]{@{}c@{}}Word Removal \\ Prediction CE Loss\end{tabular} & \begin{tabular}[c]{@{}c@{}}Word Removal \\ Prediction Focal Loss\end{tabular} \\ \toprule
% \begin{tabular}[c]{@{}c@{}}“God gave us a choice \\ and my choice is to love”\\ nnPROTECT THIS CHILD\end{tabular} & \begin{tabular}[c]{@{}c@{}}“God love”nn\\ PROTECT THIS CHILD\end{tabular} & Hope & Non\_Hope& Hope \\ \midrule
% \begin{tabular}[c]{@{}c@{}}I love how kids \\ can react so positively to this\end{tabular} & kids react positively & Hope & Non\_Hope & Hope \\ \midrule
% \begin{tabular}[c]{@{}c@{}}Good thing my country \\ never had slavery\end{tabular} & \begin{tabular}[c]{@{}c@{}}Good thing country \\ slavery\end{tabular} & Hope & Hope & Non\_Hope\\ \toprule
% \end{tabular}%
% }
% \end{minipage}
% \end{table}

Meanwhile, we can also see focal loss with $\gamma=2$ to show similar results like cross-entropy loss, except focal loss tends to add an additional 0.2\% improvement in overall results for Test data. We believe there are two significant reasons for focal failure not having any further progress in results. 
Firstly, using overlap word removal, we see that, on average, we see a reduction in sentence length by five words. As a result, examples that previously were hard to classify and required focal loss have limited context, hence having no effect with focal loss. Secondly, we can see that the 5\% of comments that are wrongly classified include a lack of correct context, as shown in Table \ref{tab:wordrem}.

Finally, in our study, we used a $\tau=25$ as a fixed 
value and did not perform any ablation study for a range of values. This is because initial experiments with a large value of $\tau=30$ show a minor decrease in results which is obtained with $\tau=25$ as the network lacks sufficient input context for learning in line with 5\% of the cases mentioned earlier. Meanwhile, a lower value $\tau=20$ also shows lower results, especially for the Hope class. This is because, initially, higher context words lead to memorizing overlapping words. However, we do believe a more in-depth analysis is needed in this regard.

To summarize, our findings are
\begin{itemize}
    \item The overall results for both validation set and test set are well improved with scores of 0.9518 and 0.9493 for cross-entropy loss compared to baseline.
    \item Meanwhile, similar behavior could be observed with focal loss, except with a minor drop in results.
    \item Large value of $\tau$ causes limitation in context information often leading to dropping in performance, meanwhile lower value of $\tau$ biases results towards Non-Hope class.
\end{itemize}

\subsection{Comparison of State of the Arts}\label{sota_res}
We compare our strong baseline with some state-of-the-art methods in Table \ref{tab:consolidated_results}. To this end, since all the existing approaches present only weighed average F1-Score, we compare our strategies against state-of-the-art using the same. However, we do explicitly highlight the macro average of F1-Score. We have divided benchmarks into three categories inline with section \ref{s1}. The consolidated influence of each of the proposed methods from section \ref{tricks} is as shown in Table \ref{tab:consolidated_results} along with a comparison of state-of-the-art methods to date. To this end, word removal is a clear winner where the net macro average is 0.17 higher than the closest state-of-the-art methods. 

\begin{table}[!htb]
\tbl{Consolidated results with different proposed strategies and comparison over state-of-the-art. Sorted in descending order of weighed average F1-Score.\textsuperscript{a}}
{
\begin{tabular}{ccc}
\toprule
\multirow{2}{*}{\textbf{}} & \multicolumn{2}{c}{\textbf{F1-Score}} \\ \cline{2-3} 
 & \textbf{Macro Average} & \textbf{Weighed Average} \\ \toprule
\textbf{M-BERT  + Focal Loss + word removal} & 0.9514 & 0.9846 \\ \cmidrule{1-3} 
\textbf{M-BERT  + word removal} & 0.9493 & 0.9839  \\ \cmidrule{1-3} 
\textbf{M-BERT + Focal Loss + Back   Translation v2} & 0.7830 & 0.9332 \\ \cmidrule{1-3} 
\citep{mahajan-etal-2021-teamuncc,huang-bai-2021-team,awatramani-2021-hopeful} & NA & 0.93 \\
\cmidrule{1-3} 
\textbf{M-BERT + Focal Loss + Back   Translation v1} & 0.7724 & 0.9294 \\
\cmidrule{1-3} 
\textbf{M-BERT + Cross Entropy Loss} & 0.6686 & 0.9290 \\\cmidrule{1-3}
\textbf{M-BERT  + Focal Loss} & 0.7859 & 0.9277 \\ \cmidrule{1-3} 
\textbf{M-BERT  + Focal Loss + Contextual Augmentation} & 0.7438 & 0.9223 \\  \cmidrule{1-3} 
\citep{zhao-tao-2021-zyj} & NA & 0.92 \\   \toprule
\end{tabular}}
\tabnote{\textsuperscript{a}We compare using Weighed Average F1 as Macro F1 is unavailable.}
\label{tab:consolidated_results}
\end{table}

% \begin{table}[!htb]
% %\begin{minipage}{320pt}
% \centering
% \caption{Consolidated results with different proposed strategies and comparison over state-of-the-art. Sorted in descending order of weighed average F1-Score.\footnote{We compare using Weighed Average F1 as Macro F1 is unavailable.}}
% \scalebox{.9}{
% \begin{tabular}{ccc}
% \toprule
% \multirow{2}{*}{\textbf{}} & \multicolumn{2}{c}{\textbf{F1-Score}} \\ \cline{2-3} 
%  & \textbf{Macro Average} & \textbf{Weighed Average} \\ \toprule
% \textbf{M-BERT  + Focal Loss + word removal} & 0.9514 & 0.9846 \\ \cmidrule{1-3} 
% \textbf{M-BERT  + word removal} & 0.9493 & 0.9839  \\ \cmidrule{1-3} 
% \textbf{M-BERT + Focal Loss + Back   Translation v2} & 0.7830 & 0.9332 \\ \cmidrule{1-3} 
% \citep{mahajan-etal-2021-teamuncc,huang-bai-2021-team,awatramani-2021-hopeful} & NA & 0.93 \\
% \cmidrule{1-3} 
% \textbf{M-BERT + Focal Loss + Back   Translation v1} & 0.7724 & 0.9294 \\
% \cmidrule{1-3} 
% \textbf{M-BERT + Cross Entropy Loss} & 0.6686 & 0.9290 \\\cmidrule{1-3}
% \textbf{M-BERT  + Focal Loss} & 0.7859 & 0.9277 \\ \cmidrule{1-3} 
% \textbf{M-BERT  + Focal Loss + Contextual Augmentation} & 0.7438 & 0.9223 \\  \cmidrule{1-3} 
% \citep{zhao-tao-2021-zyj} & NA & 0.92 \\   \toprule
% \end{tabular}}
% \label{tab:consolidated_results}
% %\end{minipage}
% \end{table}

Meanwhile, without any pre-processing, we can see that M-BERT with focal loss produces results close to works of \citep{awatramani-2021-hopeful,huang-bai-2021-team,mahajan-etal-2021-teamuncc} all of which some of the transformers. Our approach of using Focal Loss with M-BERT still outperforms results of \citep{zhao-tao-2021-zyj} which does complex processing, where it uses the attention mechanism to adjust the weight of all the output layers. Among all the published works to date on using M-BERT, our work outperforms them by a large margin, primarily upon coupling with one or more strategies to handle data imbalance. To the best of our knowledge, ours is the only model that achieves the best performance with a minor change in pre-processing and doesn't require the addition of any attention modification, local features, or semantic information. Besides, the best-performing model is away by a large margin compared to the current state-of-the-art.

\section{Conclusion and Outlook}\label{conclusion}
In this work, we studied the problem of hope speech detection by revisiting the issue of data imbalance from multiple perspectives. To this end, we introduce formulated M-BERT baseline (section \ref{base_res}) and studied focal loss, data augmentation strategies, and word removal pre-processing algorithms (section \ref{tricks}). In the due process, we find each of these strategies to improve the F1-Macro by 0.11, 0.10, and 0.17, respectively (section \ref{results}). This result is the best performance achieved by a variant of BERT without complex modifications to the best of our knowledge. We further evaluated each of these strategies in detail to find various advantages with potential drawbacks. Firstly, while we see focal loss to improve the results when coupled with M-BERT, we find $\gamma$ to be highly influential, wherewith optimal values the results are significantly enhanced (section \ref{focal_res}). However, we do notice that this also affects convergence and training time.

Meanwhile, we saw, the augmentation approaches have little impact on overall results when coupled with focal loss. But with regular M-BERT, the results seem to improve but are not comparable with Focal loss. Moreover, we find that compared to contextual augmentation, back translation to helpful F1-Macro by additional 2\% and also french language is more suitable as an intermediary for back-translation (section \ref{Aug_res}).   We also emphasized the potential bias that is introduced due to issues of word overlap in section \ref{m1}, for which we introduced a word removal algorithm. We find this method to outperform all the other strategies by a considerable margin of 0.17 in F1-Macro and also helped with both focal loss and cross-entropy loss (section \ref{removal_res}). Finally, we compared our work extensively with multiple current state-of-the-art methods.

Nevertheless, multiple aspects were unexplored, namely (a) ablation study of the effect of features from one or more layers of M-BERT, (b) convergence issues of focal loss and its relative contribution to errors in hope speech, (c) impact of the number of words augmented in contextual data augmentation, (d) relationship of intermediary language used for back translation, (e) effect of word removal on context which we plan to visit in upcoming works. In the future, we will also explore additional languages and code mixed data to verify the effectiveness of suggested strategies. In addition, some confusion remains, such as why loss function has no impact on performance when we use word removal. Points wherein the conclusions are unclear are also planned to be explored.

\section*{Acknowledgement}
Hariharan RamakrishnaIyer LekshmiAmmal was supported by the Ministry of Human Resources and Development Ph.D. Fellowship, Government of India. Bharathi Raja Chakravarthi was supported in part by a research grant from Science Foundation Ireland (SFI) under Grant Number SFI/12/RC/2289$\_$P2 (Insight$\_$2), co-funded by the European Regional Development Fund and Irish Research Council grant IRCLA/2017/129 (CARDAMOM-Comparative Deep Models of Language for Minority and Historical Languages).

Any opinions, findings, and conclusion or recommendations expressed in this material are those of the authors only and does not reflect the view of their employing organization or graduate schools. Manikandan Ravikiran did part of the work on word overlap and focal loss using language models as part of the complex assignment project in CS6460 (Spring 2020) at the OMSCS Program, Georgia Institute of Technology, which was extended for the task of hope speech detection by members of Department of Information Technology, National Institute of Technology Karnataka, Surathkal, Mangalore, India.

\section*{Conflict of Interest}
The authors have no conflicts of interest to declare that are relevant to the content of this article.

% \section*{Author Contributions}
% Hariharan, Manikandan, Anand Kumar contributed to the study conception and design. Data collection, processing and model development were performed by Hariharan, Gayathri Nisha, Navyasree, Adithya Madhusoodanan. Analysis was performed by all the authors. The first draft of the manuscript was written by Manikandan, Hairharan and Anand Kumar and reviewed by Bharathi Raja Chakravarthi. Hariharan, Manikandan, Bharathi commented and revised various sections of the manuscript. All authors read and approved the final manuscript for submission.

\section*{Code Availability}
The code is made available in \url{https://github.com/hariharanrl/lre\_hope\_2021}.

\bibliographystyle{apalike}      % basic style, author-year citations
\bibliography{ref}

\end{document}